\documentclass{article}

\PassOptionsToPackage{numbers, compress}{natbib}
\usepackage[final]{neurips_2018}

\usepackage[utf8]{inputenc} 
\usepackage[T1]{fontenc}    
\usepackage{hyperref}       
\usepackage{url}            
\usepackage{booktabs}       
\usepackage{amsfonts}       
\usepackage{nicefrac}       
\usepackage{microtype}      
\usepackage{graphicx}
\usepackage{mathabx}

\newcommand{\btr}{$\blacktriangleright$}
\newcommand{\btl}{$\blacktriangleleft$}
\newcommand{\grd}{$\nabla$}

\title{Automatic differentiation in ML: \\ Where we are and where we should be going}

\author{
  Bart van Merri\"enboer \\
  Mila, Google Brain \\
  \texttt{bartvm@google.com} \\
  \And
  Olivier Breuleux \\
  Mila \\
  \texttt{breuleuo@iro.umontreal.ca} \\
  \And
  Arnaud Bergeron \\
  Mila \\
  \texttt{bergearn@iro.umontreal.ca} \\
  \And
  Pascal Lamblin \\
  Mila, Google Brain \\
  \texttt{lamblinp@google.com} \\
}

\begin{document}



\maketitle

\begin{abstract}
  We review the current state of automatic differentiation (AD) for array programming in machine learning (ML), including the different approaches such as operator overloading (OO) and source transformation (ST) used for AD, graph-based intermediate representations for programs, and source languages. Based on these insights, we introduce a new graph-based intermediate representation (IR) which specifically aims to efficiently support fully-general AD for array programming. Unlike existing dataflow programming representations in ML frameworks, our IR naturally supports function calls, higher-order functions and recursion, making ML models easier to implement. The ability to represent closures allows us to perform AD using ST without a tape, making the resulting derivative (adjoint) program amenable to ahead-of-time optimization using tools from functional language compilers, and enabling higher-order derivatives. Lastly, we introduce a proof of concept compiler toolchain called Myia which uses a subset of Python as a front end.\@
\end{abstract}

\section{Introduction}

Recent advances in ML, and deep learning in particular, have in part been driven by advances in hardware~\cite{lecun2015deep,schmidhuber2015deep}. This increase in computational power has spurred the development of a large number of software libraries, compute kernels, programming languages, and compiler toolchains in order to exploit it. We distinguish some features and objectives that separate these ML frameworks from traditional array programming frameworks.


Firstly, many machine learning models use optimization algorithms which require access to derivatives of the model. Automatic differentiation~\cite{griewank2008evaluating} comprises a collection of techniques that can be employed to calculate the derivatives of a function specified by a computer program, and is a central feature of popular ML frameworks such as TensorFlow~\cite{abadi2016tensorflow} and PyTorch~\cite{paszke2017automatic}.


ML frameworks also put heavy emphasis on being able to iterate quickly on new models using high-level, dynamically typed languages, while maintaining high performance through aggressively exploiting resources (e.g., through parallelism, distributed computing, accelerators, static optimization). Moreover, since the derivative code is generated programmatically using AD, frameworks cannot always rely on users writing hand-tuned code and must instead provide compiler optimizations.

Despite the growth in ML frameworks, many have been developed in isolation of the AD community, and many of their insights regarding language design and interactions between source transformation and compiler optimizations have gone largely ignored. Moreover, although many ML frameworks have slowly been adopting functional language concepts (such as pure functions, immutable variables, lazy evaluation) many of the standard approaches in use by functional language compilers to guarantee high performance (A-normal form and continuation passing style representations, persistent data structures, heap recycling, etc.) have not been applied.

In some cases popular ML frameworks have sacrificed flexibility and generality compared to popular array programming packages such as NumPy~\cite{walt2011numpy} in order to provide AD and achieve high performance. On the one hand, frameworks relying on computation graphs such as TensorFlow and Theano~\cite{al2016theano} do not support higher-order functions or recursion, even though some ML models (e.g.~\cite{tai2015improved}) are more naturally expressed using recursion than loops. On the other hand, frameworks relying on operator overloading such as PyTorch and Autograd~\cite{maclaurin2015autograd} see performance degradation for models with scalars or small vectors.\footnote{\url{https://github.com/pytorch/pytorch/issues/2518}}

\section{Background and prior work}

The development of ML frameworks has been driven by a wide range of fields and perspectives---systems programming, automatic differentiation, programming languages, compiler design, applied machine learning, etc.--which has lead to duplicated research and confused terminology (e.g. \emph{define-by-run} and \emph{operator overloading}). To contextualize our proposed framework, the first half of this paper consists of a review which aims to synthesise these different perspectives. We will begin with explaining the nature of AD and the various challenges associated with it. Then we will review the different approaches to AD and relevant prior work from different domains, such as graph representations from the compiler literature, and language and IR design from functional languages. We will discuss the uses of these approaches in existing frameworks and how they affect performance, expressive power, and usability.

Given this insight, our goal is to outline in the subsequent sections a proof of concept of a high-performance ML framework with first-class support for AD, but which has the flexibility and expressive power of a generic, high-level programming language so that it does not restrict the ability of ML researchers to explore novel models and algorithms.


\subsection{Automatic differentiation}

Automatic differentiation (AD, also called algorithmic differentiation) relies on the ability to decompose a program into a series of elementary operations (primitives) for which the derivatives are known and to which the chain rule can be applied. AD allows for the calculation of derivatives of any order up to working precision.

AD has been studied since the 60s and 70s and has been employed in fields such as computational fluid dynamics, astronomy, and mathematical finance~\cite{griewank2008evaluating}. Both its implementation and its theory are still an active area of research (e.g.~\cite{siskind2016efficient} and~\cite{wang2016capitalizing}). We recommend~\cite{griewank2008evaluating} and~\cite{baydin-2018-ad-machinelearning} for a review of AD in general and in the context of machine learning respectively. From an application perspective, AD affects and interacts with the entire toolchain, from language design through intermediate representations, static analysis, to code generation and program execution.

The runtime and memory complexity of AD depends on the order in which the chain rule is evaluated. Evaluating the chain rule from right to left (from inputs to outputs) is referred to as \emph{forward mode}, whereas evaluating it from left to right (from outputs to inputs) is called \emph{reverse mode}. Forward mode has constant memory requirements and its runtime complexity scales with the number of inputs. Reverse mode's runtime complexity scales with the number of outputs, and its memory complexity grows with the number of intermediate variables. In principle, forward and reverse mode can be mixed, but finding the optimal way of doing so is NP-complete~\cite{naumann2008optimal}.

In forward mode, the partial derivatives of intermediate variables are calculated in step with the original program. As such, forward mode is relatively straightforward to implement, e.g. using dual numbers~\cite{clifford1873preliminary}. In reverse mode, the chain rule is evaluated in reverse order of the original program. This is a more complex program transformation: an \emph{adjoint} program must be constructed whose control flow is the reverse of the original (or \emph{primal}) program. First, the primal program is run to obtain the output, and then the adjoint program is run to compute the gradient, starting from that output and going backwards. In order to do so efficiently, each statement in the adjoint must have access to the intermediate variables of the original program. Hence, the AD transformation must guarantee that the intermediate variables are not destroyed or mutated.

In ML applications, large matrices of input parameters are typically updated using gradient descent on a scalar output cost. Since the number of inputs is significantly larger than the number of outputs, reverse mode AD is to be preferred. The term `backpropagation' is used to refer to the specialized application of reverse mode AD in machine learning.

Two implementation methods of AD are generally distinguished: operator overloading (OO) and source transformation (ST, also called source code transformation). Each method has its advantages and disadvantages in terms of usability, implementation, and efficiency~\cite{bischof2000computing}. We will briefly discuss them in the context of reverse mode AD.\@

\subsubsection{Operator overloading}

OO relies on a language's ability to redefine the meaning of functions and operators. All primitives are overloaded so that they additionally perform a tracing operation: The primitive is logged onto a `tape', along with its inputs to ensure that those intermediate variables are kept alive. At the end of the function's execution, this tape contains a linear trace of all the numerical operations in the program. Derivatives can be calculated by walking this tape in reverse.

The main advantage of OO is that it is straightforward to implement. Because the tracing passes through function calls and control flow, the AD logic is simplified. A significant downside is that a separate `derivative interpreter' is needed for the adjoint program. Having an embedded interpreter inside of the host language can complicate debugging and performance analysis. Moreover, since the program is traced and reversed at runtime, OO incurs overhead on each function call which can be particularly problematic if the primitives are fast to execute relative to the tracing operation. OO also does not allow for ahead-of-time optimizations on the adjoint program.

OO is the technique used by PyTorch, Autograd, and Chainer~\cite{tokui2015chainer}. Non-ML oriented AD frameworks using OO include ADOL-C~\cite{griewank1996algorithm} and CppAD~\cite{bell2012cppad}.

\subsubsection{Source transformation}

ST explicitly constructs the adjoint program. Unlike OO, ST needs to explicitly construct a program with a reversed control flow, which means that it needs transformation rules for function calls and control flow statements such as loops and conditionals. Whereas OO operates within the language, ST requires tooling such as parsers, tools to manipulate intermediate representations, and unparsers. The advantage of ST is that the AD transformation is done only once per program and hence doesn't incur overhead at runtime, which makes ST performant for a wider range of workloads. 
Moreover, the full adjoint program is available during compilation and can therefore be optimized ahead of time.

Although ST does not have to deal with the AD transformation at runtime, it must still ensure that intermediate variables from the forward pass are accessible by the adjoint. There are a variety of approaches to deal with this.

\paragraph{Tape-based} Frameworks such as ADIFOR~\cite{bischof1996adifor} and Tapenade~\cite{hascoet2013tapenade}  for Fortran and C use a global stack also called a `tape'\footnote{The tape used in ST stores only the intermediate variables, whereas the tape in OO is a program trace that stores the executed primitives as well.} to ensure that intermediate variables are kept alive. The original (primal) function is augmented so that it writes intermediate variables to the tape during the forward pass, and the adjoint program will read intermediate variables from the tape during the backward pass. More recently, tape-based ST was implemented for Python in the ML framework Tangent~\cite{van2017tangent}.

A problem of this approach is that the tape is a data structure constructed at runtime, analysis of which requires custom compiler passes~\cite{pascual2003tbr,hascoet2013tapenade}. Moreover, adjoint programs have a particular symmetric structure where intermediate variables from the first primal
statements are used by the last adjoint statements. This highly non-local structure is unsuitable for traditional compiler optimizations which act locally. Ways of addressing this interaction between AD and compiler optimizations is an ongoing research topic~\cite{siskind2016efficient,hascoet2017some}.
Finally, reading and writing to the tape need to be made differentiable in order to compute higher-order derivatives which involve multiple applications of reverse mode. For this reason most tape-based systems do not support reverse-over-reverse.

\paragraph{Closure-based}

To address some of the shortcomings of the tape-based approach, alternative approaches have been proposed which employ closures~\cite{pearlmutter2008reverse} or delimited continuations~\cite{wang2018language}. In both cases, tools from functional programming are used which can capture the environment of a statement during the forward pass, and execute the corresponding adjoint within that environment. The advantage of this approach is that no AD-specific compiler passes are needed: a functional language compiler will recognize the non-local use of the intermediate variables by the fact that they are free variables in the generated closure or continuation. This avoids the need for custom compiler passes, and allows for the application of all the tooling from functional compilers on the generated adjoint program~\cite{shivers1991control,siskind2008using}.

\subsection{Dataflow programming}

Popular ML frameworks such as Theano, TensorFlow, and MXNet~\cite{chen2015mxnet} follow the dataflow programming paradigm~\cite{johnston2004advances} and use computation graphs as their intermediate representation. These graph representations do not have scoping or recursive function calls, which means that AD is much easier to implement with ST. Since the adjoint program is part of the same dataflow graph, it can access the intermediate variables from the forward pass directly from the global scope, so neither tapes nor closures are required. Additionally, a simple liveness analysis makes it easy to keep intermediate values from the primal alive only for as long as required by the adjoint computation.

Using dataflow graphs without function calls\footnote{TensorFlow and Theano implement a type of subroutine through their \texttt{Defun} and \texttt{OpFromGraph} constructs, but these must be explicitly constructed by the user and don't support recursion.} nor scoping\footnote{TensorFlow has a concept it refers to as `scoping', but these scopes are not lexical and can be reentered at any time, so the lifetime of a value is not affected by its scope.} introduces limitations. Some of these limitations are addressed by the use of metaprogramming, but others affect the end-user (e.g., the lack of recursion and higher-order functions reduces the expressiveness of the language) and the compiler pipeline (e.g., loops cannot be represented in a principled way, which complicates their implementation).

An advantage of dataflow programming is that graphs are a natural representation for distributed computing~\cite{akidau2015dataflow}. This allows different operations to be easily distributed across different hosts, devices, and cores.

Graph-based IRs are generally useful for compilers, since the absence of an explicit ordering can simplify certain optimizations and scheduling. Theano's graph representation in particular was based on the representations used by computer algebra systems (CAS), enabling aggressive algebraic simplification and pattern matching. An SSA\footnote{Static single assignment, which essentially means each variable is assigned to exactly once.}-based graph representation~\cite{click1995simple,lindenmaier2005firm}, sometimes referred to as \emph{sea-of-nodes}, is used by the HotSpot Java compiler and the V8 TurboFan JavaScript compiler, and a graph representation using continuation-passing style (CPS, an IR commonly used in functional languages) called \emph{Thorin} also exists~\cite{leissa2015graph}.




\subsection{Programming languages and compilers}

Theano was one of the first software packages to refer to itself as a `linear algebra compiler'. Since then, more frameworks started approaching the definition and execution of ML models as a compiler problem. In the case of Theano and TensorFlow, they can be considered compilers of a custom language which must be metaprogrammed using Python as a metalanguage. The dataflow graph is an intermediate representation which is optimized using a series of compiler passes. The resulting program is compiled (e.g., XLA) and/or interpreted (e.g., the TensorFlow/Theano runtimes). Similarly, PyTorch has started optimizing its traced Python programs using just-in-time (JIT) compiler approaches.

More recently, projects such as DLVM~\cite{wei2017dlvm} and Swift for TensorFlow\footnote{\url{https://www.tensorflow.org/community/swift}} have attempted to extend existing compiler toolchains such as LLVM and Swift's intermediate language (SIL) with array programming and AD in order to create frameworks better suited for ML workflow needs.

Viewing ML frameworks as compiler toolchains raises several questions. For example, on what intermediate representations is it the easiest to apply AD and aggressive optimizations? IRs with closures as first-class objects will be able to use closure-based approaches to AD, whereas traditional SSA-based representations (such as SIL) would need to use a tape-based approach. And which IRs are most suitable for the heavy use of parallelism and distributed computing in ML?

Secondly, what should the source language be? The ML community is highly invested in Python, an interpreted, dynamically typed programming language which does not have built-in support for multidimensional arrays. More recently, frameworks have suggested using Swift (DLVM) or Julia (JuliaDiff, ~\cite{DBLP:journals/corr/RevelsLP16}), languages with static typing and built-in multidimensional arrays respectively. On the other hand, frameworks such as Theano and TensorFlow do not have an exposed source language but can only be metaprogrammed. In the AD community, there has been strong push away from traditional imperative languages such as Fortran and C to purely functional languages, since they simplify the implementation of AD and are easier to optimize. Examples of this are VLAD, a dialect of Lisp which is compiled with the Stalin$\nabla$ compiler~\cite{siskind2016efficient,pearlmutter2008reverse,siskind2008using}, DVL\footnote{\url{https://github.com/axch/dysvunctional-language}}, and DiffSharp~\cite{baydin2016diffsharp}.

\subsubsection{Python}

Because Python plays an important role in the ML community many popular ML frameworks are Python-based. However, the language's characteristics make it difficult to implement a high-performance AD-enabled ML framework in Python directly. The reference implementation of Python, CPython, has effectively no support for concurrency, and the interpreter is relatively slow. Moreover, its highly dynamic nature makes source transformation difficult (Tangent imposes several restrictions on the use of Python in order for it to perform ST). Python does not have built-in support for multidimensional arrays, which are only supported through third-party frameworks such as NumPy. 

How to reconcile users' desire to work in Python because of its flexibility with the need for high performance and speed is an open question. ML frameworks have focused on metaprogramming and using C extensions, but other approaches are possible. For example, Cython~\cite{behnel2011cython} is a superset of Python which compiles to Python modules, whereas Numba~\cite{lam2015numba} can compile individual Python functions using LLVM.

\section{Graph-based direct intermediate representation}

We endeavor to combine several of the aforementioned techniques and insights from the compiler and AD literature in order to provide a flexible basis for an ML framework. This requires a well-tailored intermediate representation which avoids the pitfalls of previous methods, while keeping their strengths. Concretely, we propose an IR with the following properties:

\paragraph{Graph based}

Similar to Theano or TensorFlow, programs are represented as graphs. Graphs have the advantage of being easy to optimize and flexible about execution order, as operations that do not depend on each other in the graph may be executed in any order, or in parallel. Unlike Theano and TensorFlow, however, functions may be called recursively and they are first-class objects. Functions may be passed as parameters to other functions, or returned from a function and then called. A large variety of control flow constructs, ranging from simple loops to graph traversals, can be implemented using these capabilities. Other graph frameworks tend to implement only a few of these as specialized operators, such as Theano's \texttt{scan} or TensorFlow's \texttt{while}, leading to an IR which is both more complex and less powerful than the general one we are proposing. A general IR does require more work to transform and optimize in a provably correct way in the context of automatic differentiation, but this work only needs to be done once.

\paragraph{Purely functional}

Mutation and side effects are problematic for reverse mode AD, where the backward pass requires access to the unchanged intermediate variables from the forward pass. They also interact poorly with complex optimizations because of aliasing. Restricting our language to be purely functional therefore allows us to implement more robust AD and more advanced optimizations compared to imperative languages.

Note that Myia's intended use case is not the writing of efficient low-level kernels, which often requires fine-grained memory control. Similarly to, e.g., TensorFlow, the user can write efficient low-level kernels and their derivatives in a low-level language such as CUDA or XLA, and expose them to Myia as primitives.

\paragraph{Closure representation}

AD on functional languages involves storing the primal's intermediate results into closures which are then connected together to form the adjoint. It is therefore important to have a natural representation for closures. As in Thorin, we represent a function's graph's free variables as direct pointers to nodes that belong to other graphs, thereby creating an implicit nesting relationship between them (a graph $G_c$ is ``nested'' in $G_p$ if it points to a node in $G_p$, or to a graph nested in $G_p$, or to a node in a graph nested in $G_p$). This facilitates joint optimization of a closure with the functions it is nested in. Closures are also a great means for abstraction and a natural way to represent the methods of objects, so there is a concrete advantage in expressiveness from the user's perspective, which cannot be found in other frameworks.

\paragraph{Strongly typed}

In its canonical form, every node must be associated with a concrete type. This is important to maximize performance. This is also important in ML applications, because operations tend to be very costly and it is best to catch errors as early as possible. In addition to data types, there is also a need to infer other properties such as the dimensions of vectors and matrices so that we can guarantee that the inputs of all operations have compatible dimensions prior to executing them. Type and shape inference are more complex and powerful on our proposed IR than in dataflow graphs because of the need to support recursive calls and higher order functions.

\subsection{IR specification}

\begin{figure}
    \centering
    \includegraphics[width=1.0\textwidth]{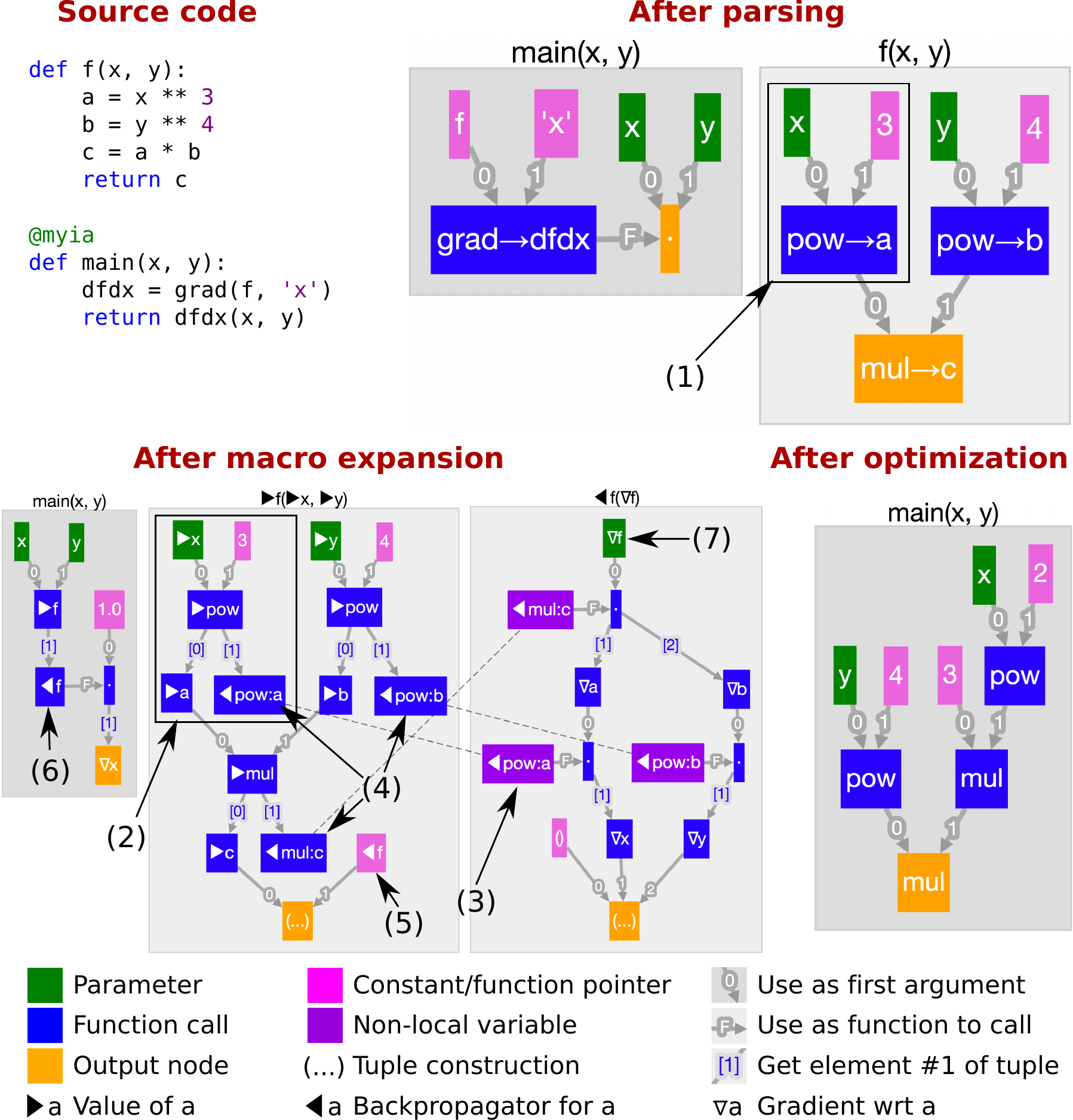}
    \caption{Transform of a simple Python program into Myia's representation. The program computes the gradient of \texttt{f} with respect to \texttt{x}. (1) identifies the part of the graph that implements \texttt{x ** 3}, or \texttt{pow(x, 3)}. After the \texttt{grad} macro is expanded, a new graph, \texttt{\btr f} is built. In that graph, \texttt{a = pow(x, 3)} becomes \texttt{\btr a, \btl pow = \btr pow(\btr x, 3)} (2). The \texttt{\btr pow} operation thus returns two values instead of one, \texttt{\btr a} being equal to the original value \texttt{a}, and \texttt{\btl pow} being the \emph{backpropagator} for this operation. That backpropagator is used in (3). It is applied on the gradient wrt the output, \texttt{\grd a}, and produces the gradient wrt the input \texttt{\grd x} (it also produces a gradient wrt the constant \texttt{3}, but that gradient is not used). (4) points to all the backpropagators created while executing \texttt{\btr f}. These are closures that retain pointers to all the information necessary to perform reverse mode AD. \texttt{\btl f}, the backpropagator we construct for \texttt{f}, is returned by \texttt{\btr f} (5), alongside the main result (notice that this mirrors the interface in (2)). \texttt{\btl f} retains pointers to all backpropagators in (4), as indicated by the dashed lines. \texttt{\btl f} is retrieved in (6), and immediately called with the value \texttt{1.0} for the parameter \texttt{\grd f} (7). In this context, \texttt{\grd f} corresponds to $\partial f / \partial f$, hence why we give it a value of one. After optimization, all functions and backpropagators end up being inlined. All unused computations are cut, and what remains is an expression for $\partial f / \partial x$ that is essentially identical to what one would have written by hand.}
    \label{fig:pipeline}
\end{figure}

Concretely, our representation represents a function as a graph object with a list of parameter nodes and a single return node (multiple return values are supported through tuples). A node represents a function application and has an ordered list of incoming edges. The first incoming edge is a pointer to the function to apply, and the rest point to the arguments. Constants are represented as nodes with no incoming edges and a value field. Links between nodes are bidirectional, so that graphs can be traversed in either direction. Each non-constant node belongs to a single graph. See Figure \ref{fig:pipeline} for a visual representation of the IR.

Compared to other representations, our representation is more expressive than dataflow graphs, and more flexible than SSA or CPS representations which tend to be rigid about execution order. It is closest to A-normal form (ANF,~\cite{flanagan1993essence}), where every intermediate computation is assigned a unique name, but it is graphical rather than syntactic and therefore easier to manipulate algorithmically.

\subsection{Source transformation}

\label{sec:st}

AD can be implemented for this IR using ST with a closure-based method. We closely follow the approach described in~\cite{pearlmutter2008reverse}. The transformed program constructs a chain of closures during the forward computation. These closures contain the adjoint code required to compute the derivatives along with the intermediate variables from the forward pass that are needed.

The transformation proceeds as follows: Each function call is transformed to return an additional value, which is a closure called the `backpropagator'. The backpropagator computes the derivative with respect to the inputs given the derivatives with respect to the outputs. The backpropagators of primitives are known, whereas the backpropagators of user-defined functions can be easily constructed by calling the backpropagators of the function calls in the body in reverse order.

In order to ensure that our transformation can be applied again on the transformed program (so we can use reverse-over-reverse to compute second-order derivatives), it must be able to handle functions with free variables. To this end, each backpropagator will return the partial derivatives with respect to the inputs of the original function, as well as an ordered set of partial derivatives with respect to the free variables. The backpropagator of the function that built the closure is responsible for unpacking these partial derivatives so that it can add contributions to the free variables that belong to it, this unpacking being the adjoint of closure creation. Closures are first class functions: when given as inputs of other closures, they are treated like any other input.

\section{Myia}

Myia is a functioning proof of concept of a toolchain that uses the proposed graph representation\footnote{Code available at \url{https://github.com/mila-udem/myia}}. Myia performs type inference given the input types, and applies a series of optimizations such as inlining, common expression elimination, constant propagation, closure conversion, and algebraic simplifications. The final code can be executed using an interpreter, and we also implemented a prototype which compiles the straight-line parts of the graph using TVM~\cite{chen2018tvm}.

\subsection{Python front end}

Due to Python's popularity in the ML community, we feel it is important to offer a front end in that language. Users can write models in a subset of Python 3.6 and have them compiled to our IR. This requirement is ostensibly at odds with our IR being pure and strongly typed, for Python is neither of these things. We solve that apparent contradiction by selecting a pure subset of Python, and running an advanced type inference algorithm on the functions the user asks to compile.
In that sense, our approach is similar to that of Numba and Cython, or the recently introduced \texttt{@script} decorator in PyTorch\footnote{\url{https://pytorch.org/2018/05/02/road-to-1.0.html}}. Functions that should be compiled with Myia are denoted using the \texttt{@myia} decorator, and can be freely mixed with Python code in the same file.

Most of Python's features, such as functions, conditionals, and loops, can readily be parsed into our functional representation. However, Python does include some statements such as index assignment (\texttt{x[i] = v}) and augmented assignment statements (\texttt{x += y}) which imply mutability. We currently forbid these statements in Myia, although it may be possible to support principled use of them in the future through techniques like uniqueness typing~\cite{barendsen1993,devries2008}.

Myia uses Python's \texttt{inspect} module to parse the function into an abstract syntax tree (AST),
and converts that AST into the graph representation we previously described. Source transformation as described in Section~\ref{sec:st} is used to generate the code for derivatives. 
See Figure~\ref{fig:pipeline} for an illustration of how a Python function is parsed into the proposed IR, its adjoint program is created using ST, and finally optimized to produce an efficient derivative function.




\subsection{Type inference}


Python is a dynamically typed language, but for the sake of optimization and eager error reporting, it is important to be able to infer concrete types for all expressions. While it is possible to write optional type annotations in Python 3.6, they are not widely used in practice, and we wish to minimize the amount of work one has to do in order to port existing code to Myia.

When a Myia function is called, we use the types of the user-provided arguments as a starting point for type inference, which allows us to compile a specialized version of the function for these types. No type annotations are required, even when using higher order functions such as \texttt{map} or \texttt{grad}. Myia functions can be polymorphic: Myia will specialize each use of a function according to the input type signature for that call site. This means users can write highly dynamic programs just as they are used to in Python, and Myia will check them.

The inferrer operates on an untyped version of the IR. It can infer types as well as values (constant propagation) and shapes. Inference for other properties can easily be added in the future. The inferrer is implemented using coroutines: to infer a certain property through a certain primitive, one may write a coroutine (\texttt{async def} in Python) that asynchronously requests any number of properties from any number of nodes and combines the results using arbitrary logic.




\subsection{Optimization}


Reverse mode AD in Myia poses a few specific challenges for optimization that we have to tackle. As may be seen in Figure \ref{fig:pipeline}, the AD transform produces graphs that are substantially larger than the original source. These graphs typically contain many computations that are not necessary, such as gradients with respect to constants, and a lot of tuple packing and unpacking. These graphs can be simplified using inlining and local optimizations. Figure \ref{fig:pipeline} demonstrates the resulting simplification.

\section{Conclusion}

In this work we examined the different approaches and techniques used in developing AD-enabled ML frameworks, drawing insights from functional languages, graph-based IRs, and AD. To address some of the shortcomings in existing frameworks, we propose a novel graph-based intermediate representation and describe a proof of concept toolchain called Myia to show its advantages.

The result is a system that can achieve performance similar to compiled frameworks such as TensorFlow, while providing the flexibility of OO frameworks such as PyTorch with e.g.\ support for recursion and higher-order functions.

We believe that as AD frameworks will slowly move towards being full-fledged languages and compilers, developers will benefit from building on many other ideas from these fields. For example, other techniques from functional languages that could be beneficial include the use of monads to handle random number generators, and using higher-order functions for kernel programming (similar to Tensor Comprehensions\footnote{\url{https://facebookresearch.github.io/TensorComprehensions/}}).

\subsubsection*{Author contributions and acknowledgements}

Bart van Merri\"enboer worked on the design and implementation of the IR, as well as the design of the AD system. Olivier Breuleux worked on the design of the IR and the type system, and on the implementation of the IR, type inference, the AD system, and optimization and compiler pipeline. Arnaud Bergeron worked on shape inference, on Myia's virtual machine, and integrating the GPU backend. Pascal Lamblin worked on the design of the AD system and the organization of the project.

The authors would like to thank Maxime Chevalier-Boisvert for her work on an earlier design and prototype. Her contributions and insight helped shape the current version of Myia. Early discussions and brainstorming with Olexa Bilaniuk also helped determining the scope and direction of the project.

\bibliographystyle{plainnat}
\bibliography{myia}

\end{document}